\documentclass[conference]{IEEEtran}
\ifCLASSINFOpdf

\else

\fi

\usepackage{amsmath} % assumes amsmath package installed
\usepackage{mathtools}
\usepackage{bm}
\usepackage{multirow}
\usepackage{tikz,pgfplots}
\usetikzlibrary{arrows.meta,bending,positioning}
\usepackage{tikz-dimline}
\usepackage{subcaption}
\usepackage{textcomp}
\begin{document}

\title{TO-mdiSPAs: Topology Optimization of multi-directional Soft Pneumatic Actuators}

% author names and affiliations
% use a multiple column layout for up to three different
% affiliations
\author{\IEEEauthorblockN{Swagatam Islam Sarkar}
\IEEEauthorblockA{Department of Mechanical and Aerospace Engineering\\
Indian Institute of Technology Hyderabad\\
Telangana-502284, India\\
Email: swagatam1992@gmail.com}
\and
\IEEEauthorblockN{Prabhat Kumar}
\IEEEauthorblockA{Department of Mechanical and Aerospace Engineering\\
Indian Institute of Technology Hyderabad\\
Telangana-502284, India\\
Email: pkumar@mae.iith.ac.in}

}

\maketitle

\begin{abstract}
Soft pneumatic actuators (SPAs) are highly promising and have been extensively explored within the field of soft robotics. Under pneumatic pressure loads, an SPA undergoes bending deformation to perform specific tasks. A multi- or omni-directional SPA can bend and move in any direction within a 3D space, leveraging its multiple degrees of freedom to realize various mechanical functions. This work presents a systematic methodology using topology optimization (TO) to achieve an optimized design for a multi-directional SPAs. To ensure manufacturing robustness, a three-field TO formulation considering blueprint, dilated, and eroded designs is implemented. Additionally, Darcy's law, incorporating a drainage term, is used to model the design-dependent nature of the pneumatic loading. A min-max optimization problem is formulated based on the target output deformations of the SPA unit and solved using the Method of Moving Asymptotes. The optimization yields a high-performing, unconventional geometric design. Finally, numerical simulations demonstrate that the optimized SPA successfully achieves versatile multi-directional movements.
\end{abstract}

\IEEEpeerreviewmaketitle

\section{INTRODUCTION}

Robots are primarily designed to minimize human intervention in performing specific tasks. The flexibility of conventional robotic systems remains constrained by their reliance on rigid linkages and chain mechanisms~\cite{xavier2022soft,kumar2022towards}. In contrast, soft robots offer superior flexibility and dexterity, as they are fabricated from compliant materials with Young's modulus values typically in the kilopascal to megapascal range~\cite{xavier2022soft,kumar2022towards}. These are categorized based on their actuation types as fluidic pressure-driven (pneumatic) actuation~\cite{kumar2022towards,de2020topology,kumar2022topological,kumar2024sorotop}, shape-memory material-based actuation~\cite{jin2016soft}, cable-driven actuation~\cite{chen2018topology}, dielectric elastomer actuation~\cite{pourazadi2019investigation}, etc. Among these, fluidic pressure–driven (pneumatic) robots are widely preferred owing to their lightweight structure, rapid actuation response, and utilization of inexpensive components~\cite{kumar2022towards}. These soft robots are called soft pneumatic actuators (SPAs). They are classified into four categories based on their actuation mechanisms and motion characteristics: extending/contracting, bending, helical/twisting, and bidirectional/omnidirectional actuators.

The McKibben actuators~\cite{zhang2021fluid,tondu2012modelling,zhang2019robotic} represented one of the earliest works of pneumatic actuation for soft robots. Building upon such concepts, Drotman et al.~\cite{drotman20173d} designed and fabricated a soft actuator capable of achieving biaxial rotational motion. Inspired by biological systems, Xiao et al.~\cite{xiao2021modeling} introduced a multi-degree-of-freedom pneumatic muscle, modeled after leeches, capable of bending in multiple directions and extending its soft body.
Suzumori et al.~\cite{suzumori1991flexible,suzumori1991development} proposed multidirectional actuators that have a robotic arm with many degrees of freedom and can move like snakes. Yan et al.~\cite{yan2016three} presented the design and fabrication of a pneumatic-powered soft actuator module to achieve omnidirectional bending motion. Xie et al.~\cite{xie20183d} attained extending and omnidirectional motion by a three-chambered soft actuator made of high-resistance rubber. 
Skorina et al.~\cite{skorina2018reverse} developed an analytical model of a reverse pneumatic artificial muscle and proposed a dynamic motion control strategy to ensure precise bidirectional position tracking under step inputs. Liu et al.~\cite{liu2019simulation} proposed an SPA capable of bending within two vertical planes. Chen et al.~\cite{chen2019fabrication} designed and fabricated a bidirectional bending SPA integrated with an embedded curvature sensor.

The SPAs mentioned above are primarily based on heuristic design principles that rely on the designer's prior experience and intuition. Despite the availability of various SPA configurations, a systematic design methodology remains limited. In addition, 
Although advancements in the output motion of soft robots have improved their performance, their actuation remains limited to planar motion. Furthermore, most soft robotic applications demand complex maneuvers and task execution in three-dimensional space.
To address these gaps, the present work introduces a systematic design of multidirectional soft pneumatic actuators (mdiSPAs) using a topology optimization (TO) framework that explicitly considers the design-dependent characteristics inherent to pneumatic actuation. A significant advantage of TO lies in its ability to automatically generate optimal geometries without prior knowledge of the final topology.

A soft pneumatic actuator functions similarly to a compliant mechanism (CM)~\cite{kumar2022towards,kumar2026soft}. Both achieve motion through the deformation of flexible structures. Therefore, with suitable modifications, TO approaches developed for CMs under pressure loads can be effectively adapted to optimize SPA designs~\cite{kumar2022towards,kumar2022topological,kumar2024sorotop}. TO determines the optimal material distribution within a defined design domain to achieve specified performance objectives. This approach is widely used in engineering and product design to develop mechanically efficient, lightweight structures. The design domain is discretized into a finite number of selected elements~\cite{kumar2023honeytop90}, with each element assigned a design variable \(\rho \) \((0 \leq \rho \leq 1)\), where \(\rho = 1\) represents a solid region and \(\rho = 0\) denotes a void region. In pneumatic systems, the applied load is design-dependent, i.e., its location, direction, and magnitude evolve during the optimization process~\cite{kumar2020topology}. This dependency introduces significant challenges within a TO framework~\cite{kumar2020topology,kumar2023topress}. 

Only a handful of studies address the optimization of pneumatically actuated CMs using TO, for example, see Refs.~\cite{de2020topology,kumar2022topological,kumar2024sorotop,kumar2020topology,lu2021topology,kumar2021topology}. Kumar et al.~\cite{kumar2020topology} proposed a method based on Darcy's law coupled with a drainage quantity to model pneumatic actuation within a TO setting. Kumar~\cite{kumar2022towards} used the method proposed in~\cite{kumar2020topology} for designing soft bending robots that account for the design-dependent behavior of pneumatic loads. Kumar and Langelaar~\cite{kumar2022topological} and Pinskier et al.~\cite{pinskier2024diversity} applied the Darcy method~\cite{kumar2020topology} to design robust CMs and multi-material soft pneumatic grippers, respectively. Chi and Liu~\cite{chi2025topology} used a TO framework for SPAs subjected to design-dependent pressure loads. Liu et al.~\cite{liu2018optimal} proposed a TO strategy for designing a two-finger underactuated soft robotic gripper with compliant, additively manufactured fingers. Zhang et al.~\cite{zhang2018design} established an integrated TO framework that couples MATLAB and Abaqus to automate the design. Chen et al.~\cite{chen2019optimal} developed a Bi-directional Evolutionary Structural Optimization-based framework for the automatic design of soft pneumatic bending actuators. Kumar et al.~\cite{kumar2026soft} employed TO to design PneuNet-based soft pneumatic grippers. They experimentally demonstrated the gripper's performance by grasping various objects under different pressure loads. To design such actuators, finite strain deformation is considered in~\cite{mehta2026topology,dalklint2026shape}. Sarkar and Kumar~\cite{sarkar2025TO} developed a bi-directional SPA capable of producing motion in two opposite directions under pneumatic pressure loading. Building upon this concept, the present work generalizes the actuation mechanism to achieve motion in arbitrary directions by appropriately varying the magnitudes and application locations of the pressure loads. A robust formulation extensively avoids manufacturing tolerances. The worst-case scenario is over-etching and an eroded design; the actual design is the blueprint. In this work, we adopt a Darcy-based method to address the design-dependent behavior of pneumatic pressure loads within a density-based TO framework for the systematic design of \textit{mdiSPAs} using a robust approach.

The rest of the paper includes the modeling of pressure loads in Sec.~\ref{sec2}, the formulation of the optimization problem in Sec.~\ref{sec3}, the results and discussion in Sec.~\ref{sec4}, and the conclusions in Sec.~\ref{sec5}.

\section{Modeling of pressure-induced pneumatic loading} \label{sec2}

This section provides a concise overview of the method for modeling pneumatic pressure loads based on Darcy's law. Additional theoretical and implementation details can be found in~\cite{kumar2020topology,kumar2023topress,kumar2021topology}.

During the optimization process, the design continuously evolves, resulting in a finite element (FE) model comprising both solid and void elements. To accurately represent the applied pneumatic load under these evolving material states, Darcy's law is employed. In this formulation, the flux $q$ is expressed as~\cite{kumar2020topology,kumar2023topress,kumar2021topology}:
\begin{equation}
    q=-\frac{k}{\mu}\nabla p,
    \label{Darcy}
\end{equation}
where $k$ denotes the permeability of the medium, $\mu$ represents the fluid viscosity, and $\nabla p$ is the pressure gradient. The ratio $\frac{k}{\mu}$, referred to as the flow coefficient, is denoted by $K(\bar{\rho})$. Here, $\bar{\rho}$ denotes the physical design variable. For a given finite element $e$, the flow coefficient $K(\bar{\rho})$ is defined as
\cite{kumar2020topology,kumar2023topress,kumar2021topology}:
\begin{equation}
    K(\bar{\rho}_e)=K_v \left(1-(1-\epsilon)H(\bar{\rho}_e,\beta_k,\eta_k)\right).
\end{equation}
A drainage term, given by $Q_{drain} = -D(\bar{\rho}_e)(p - p_{ext})$, is incorporated into the model to induce localized pressure gradients near the solid–void interfaces~\cite{kumar2020topology,kumar2023topress,kumar2021topology}. Here, $p$ and $p_{ext}$ denote the instantaneous and external pressures, respectively, while $D(\bar{\rho}_e)$ represents the drainage coefficient~\cite{kumar2020topology,kumar2023topress,kumar2021topology}.  
Using Eq.~\ref{Darcy} together with the drainage term, the final equilibrium equation can be expressed as
\cite{kumar2020topology,kumar2023topress,kumar2021topology}:
\begin{equation}
    \nabla \cdot q - Q_{drain}=0.
    \label{balance1}
\end{equation}
By discretizing the structure into finite elements and performing the standard assembly procedure, Eq.~\ref{balance1} can be reformulated as follows:
\begin{equation}
    \mathbf{A} \mathbf{P} = \mathbf{0},
    \label{balance3}
\end{equation}
where the global flow matrix and the global pressure field are represented by $\mathbf{A}$ and $\mathbf{P}$, respectively. The resulting pressure field yields the nodal forces that, in turn, drive the structural deformation. This relationship can be expressed as~\cite{kumar2020topology,kumar2023topress,kumar2021topology}:
\begin{equation}
    \mathbf{F} = \mathbf{K} \mathbf{U} = -\mathbf{TP},
\end{equation}
where the global force vector is denoted by $\mathbf{F}$, while $\mathbf{K}$ and $\mathbf{U}$ represent the global stiffness matrix and the displacement vector, respectively. The matrix $\mathbf{T}$ corresponds to the global transformation matrix~\cite{kumar2020topology,kumar2023topress,kumar2021topology}.  

The next section outlines the TO formulation developed for the design and optimization of an mdiSPA.

\section{Topology Optimization of SPA} \label{sec3}

The density-based TO framework is used for solving this optimization problem. Material property (Young's modulus) is interpolated using the modified SIMP approach as \cite{sigmund2007morphology}:
\begin{equation}
	E_e = E_\text{v}+(\bar{\rho}_e)^p(E_\text{s}-E_\text{v}),
    \label{Material_interpolation}
\end{equation}
where $E_e$ is the interpolated Young's modulus of an element~$e$, $p$ is a penalization factor ($p=3$). $E_\text{s}$ and $E_\text{v}$ are the Young's modulus of solid and void material, respectively. $E_\text{v}$ is a very small value assigned to the void elements to avoid singularity in the stiffness matrix. The Heaviside projection filter is used to evaluate $\bar{\rho}_e$ (the physical design variable of element $e$), in terms of filtered design variable $\tilde{\rho}_e$ \cite{sigmund2007morphology}.

\subsection{multidirectional SPA}

\begin{figure}
    \centering
    \begin{tikzpicture}[scale=0.25]
        \fill [gray!50] (0,0) rectangle (4,8);
        \fill [gray!50] (5,0) rectangle (9,8);
        \fill [gray!50] (15,0) rectangle (19,8);
        \fill [gray!50] (20,0) rectangle (24,8);
        \fill [gray!50] (0,0) rectangle (-1,2);

        \fill [gray!50] (4,0) rectangle (5,2);
        \fill [gray!50] (9,0) rectangle (10,2);
        \fill [gray!50] (14,0) rectangle (15,2);
        \fill [gray!50] (19,0) rectangle (20,2);

        \draw[dashed, thick] (-1,1) -- (1,1);
        \draw[dashed, thick] (1,1) -- (1,7);
        \draw[dashed, thick] (1,7) -- (3,7);
        \draw[dashed, thick] (3,7) -- (3,1);
        \draw[dashed, thick] (3,1) -- (6,1);
        \draw[dashed, thick] (6,1) -- (6,7);
        \draw[dashed, thick] (6,7) -- (8,7);
        \draw[dashed, thick] (8,7) -- (8,1);
        \draw[dashed, thick] (8,1) -- (10,1);
        \filldraw [black] (11,2) circle (3pt);
        \filldraw [black] (12,2) circle (3pt);
        \filldraw [black] (13,2) circle (3pt);
        \filldraw [black] (11,1) circle (3pt);
        \filldraw [black] (12,1) circle (3pt);
        \filldraw [black] (13,1) circle (3pt);
        \draw[dashed, thick] (14,1) -- (16,1);
        \draw[dashed, thick] (16,1) -- (16,7);
        \draw[dashed, thick] (16,7) -- (18,7);
        \draw[dashed, thick] (18,7) -- (18,1);
        \draw[dashed, thick] (18,1) -- (21,1);
        \draw[dashed, thick] (21,1) -- (21,7);
        \draw[dashed, thick] (21,7) -- (23,7);
        \draw[dashed, thick] (23,7) -- (23,1);
        \draw[dashed, thick] (23,1) -- (24,1);
        \fill [black] (-1,0) rectangle (24,-2);

        \draw[black, very thick] (-1,2) -- (0,2);
        \draw[black, very thick] (0,2) -- (0,8);
        \draw[black, very thick] (0,8) -- (4,8);
        \draw[black, very thick] (4,8) -- (4,2);
        \draw[black, very thick] (4,2) -- (5,2);
        \draw[black, very thick] (-1,0) -- (24,0);            
        \draw[black, very thick] (5,2) -- (5,8);
        \draw[black, very thick] (5,8) -- (9,8);
        \draw[black, very thick] (9,8) -- (9,2);
        \draw[black, very thick] (9,2) -- (10,2);
        \draw[black, very thick] (14,2) -- (15,2);
        \draw[black, very thick] (15,2) -- (15,8);
        \draw[black, very thick] (15,8) -- (19,8);
        \draw[black, very thick] (19,8) -- (19,2);
        \draw[black, very thick] (19,2) -- (20,2);
        \draw[black, very thick] (20,2) -- (20,8);
        \draw[black, very thick] (20,8) -- (24,8);
        \draw[black, very thick] (24,8) -- (24,0);

        \draw[black, very thick] (24,0) -- (24,-2);

        \draw[cyan!50, very thick, -Stealth] (-2,0.5) -- (2,0.5);
        \node at (-2,1.4) {$p$};
        \draw[red, very thick, -Stealth] (22,8) -- (22,6);
        \node at (22,9) {$u_\text{o}$};
        \filldraw [black] (22,8) circle (4pt);
        
    \end{tikzpicture}
    \caption{Cross section of a symmetric quarter of an arm with multiple chambers}
    \label{full_SPA}
\end{figure}
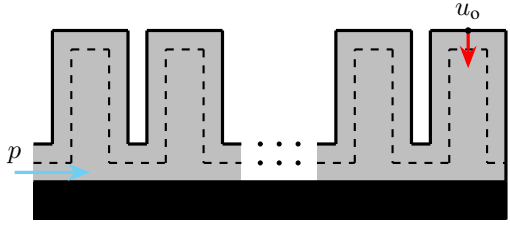

\begin{figure}
    \centering
            \begin{tikzpicture}[scale=0.5]
                \fill [gray!50] (0,0) rectangle (2,8);
                \fill [white] (1.8,0) rectangle (2,7.2);
                \fill [black] (0,0) rectangle (0.2,0.4);
                \draw[black, very thick] (0,0) -- (2,0);
                \draw[black, very thick] (2,0) -- (2,8);
                \draw[black, very thick] (2,8) -- (0,8);
                \draw[black, very thick] (0,8) -- (0,0);
                \dimline[color=brown,extension start length=-12pt, extension end length=-12pt,label style={fill=none, yshift=-6pt}]{(0.2,-2.5)}{(2,-2.5)}{\scriptsize $0.9L_x$};
                \dimline[color=brown,extension start length=-12pt, extension end length=-12pt,label style={midway,      anchor=center,fill=white,yshift=0pt}]{(0,-3.75)}{(2,-3.75)}{\scriptsize $L_x$};
                \dimline[color=brown,extension start length=-12pt, extension end length=-12pt,label style={fill=none, yshift=-6pt}]{(0,-1.5)}{(1.8,-1.5)}{\scriptsize $0.9L_x$};
                \dimline[color=brown,extension start length=12pt, extension end length=12pt,label style={midway,      anchor=center,fill=white, yshift=0pt}]{(-1.5,0)}{(-1.5,8)}{\scriptsize $L_y$};
                \dimline[color=brown,extension start length=12pt, extension end length=12pt,label style={midway,      anchor=center,fill=white, yshift=0pt}]{(-0.6,0.4)}{(-0.6,8)}{\scriptsize $19L_y/20$};
                \dimline[color=brown,extension start length=-12pt,    extension end length=-12pt,label style={midway,      anchor=center,fill=white,yshift=0pt}]{(4.6,0)}{(4.6,7.2)}{\scriptsize $0.9L_y$};
        
                \draw[cyan!50, very thick, -Stealth] (3.2,0) -- (2,0);
                \draw[cyan!50, very thick, -Stealth] (3.2,1) -- (2,1);
                \draw[cyan!50, very thick, -Stealth] (3.2,2) -- (2,2);
                \draw[cyan!50, very thick, -Stealth] (3.2,3) -- (2,3);
                \draw[cyan!50, very thick, -Stealth] (3.2,4) -- (2,4);
                \draw[cyan!50, very thick, -Stealth] (3.2,5) -- (2,5);
                \draw[cyan!50, very thick, -Stealth] (3.2,6) -- (2,6);
                \draw[cyan!50, very thick, -Stealth] (3.2,7) -- (2,7);
                \draw[cyan!50, very thick, -Stealth] (3.2,8) -- (2,8);
        
                \draw[cyan!50, very thick, -Stealth] (0,-1) -- (0,0);
                \draw[cyan!50, very thick, -Stealth] (1,-1) -- (1,0);
                \draw[cyan!50, very thick, -Stealth] (2,-1) -- (2,0);
    
                \draw[decoration={aspect=0.3, segment length=1mm, amplitude=1mm,coil},decorate] (2,9.5) -- (2,8);
                \fill [black] (1.6,9.5) rectangle (2.4,9.7);
                \draw[violet, thick, -Stealth] (2,9) -- (2,8);
                \filldraw [black] (2,8) circle (4pt);
                       
                \node at (2.5,-1) {$p$};
                \node at (3.8,3.6) {$p$};
                % \node at (2.4,8.4) {$1$};
                \node at (1.3,9) {$k_{\text{o}}$};
                \node at (2.6,8.5) {$u_{\text{o}}$};
            \end{tikzpicture}
        \caption{Design domain of one chamber (symmetric half)}
        \label{half_SPA}
\end{figure}
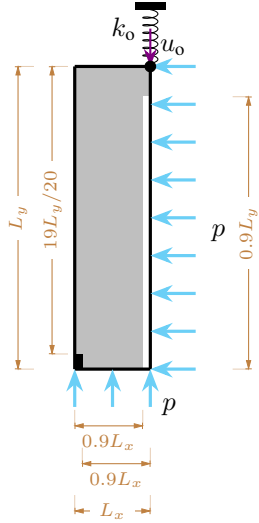

Figure~\ref{full_SPA} illustrates the design domain of the symmetric quarter of an arm with multiple chambers. The circular dots at the middle of the figure indicate that it has more than four chambers; in the actual case, we use ten chambers. The black region at the bottom of the figure corresponds to the non-design domain. The red arrow indicates the desired deformation ($u_\text{o}$) direction and the light blue arrow represents the direction of pneumatic load $p$, through the flow channel.

Figure~\ref{half_SPA} presents the symmetric half section of a single chamber subjected to pressure loading conditions. The displacement ($u_{\text{o}}$) at the output location is indicated by the violet arrow. A linear spring with stiffness $k_{\text{o}}$ is employed to represent the stiffness of the interacting workpiece. The light blue arrows indicate the regions where pressure loads are applied. The small black region at the lower left and the white region along the right edge are treated as non-design domains, as shown in Fig.~\ref{half_SPA}.

To get the output motion in multiple directions, we place the chambers in four directions, i.e., front (\textbf{\textit{f}}), back (\textbf{\textit{b}}), left (\textbf{\textit{l}}), right (\textbf{\textit{r}}) (see Fig.~\ref{optimized_SPA_3dview}). Four rectangular ducts or channels ($C_f,C_b,C_l,C_r$) are provided in four directions correspondingly, to apply the pneumatic pressure load. As per the design, when the front channel ($C_f$) is activated, and the others are not, the entire arm moves towards the rear; similarly, for the other channels. The idea is that we activate more channels together and may apply the same or different input pressures to them, enabling us to achieve output movement in any direction, depending on the positions and magnitudes of the activated channels' pressure loads. For instance, applying equal pressure loads to the $C_f$ and $C_r$ channels causes the SPA to move along a $45$\textdegree trajectory toward the back-left, which corresponds to the vector sum of the two directional movements.

\subsection{TO formulation}
To obtain the optimized chamber, a robust TO formulation is employed which is expressed as~\cite{kumar2024sorotop}:
\begin{equation}
	\begin{rcases}
		\begin{aligned}
			&{\min_{\bm{\bar{\rho}}(\bm{\tilde{\rho}(\bm{\rho})})}} :\quad \max(\mathbf{L}^T\mathbf{U}_b,\,\mathbf{L}^T\mathbf{U}_e)\\
			&\text{subjected to:}\\
			&\bm{\lambda}_{1r}:\,\,\mathbf{A}_r \mathbf{P}_r = \mathbf{0}; \quad r = b,\,e\\
            &\bm{\lambda}_{2r}:\,\,\mathbf{K}_r \mathbf{U}_r = \mathbf{F}_r = -\mathbf{T} \mathbf{P}_r\\
			&\Lambda_{b}:\frac{V_b}{V^*}-1 \le 0 \\
            &\Lambda_{e}:\frac{S_e}{{S}^*}-1 \le 0 \\
			&\quad\,\,\,\, 0 \leq \rho_{ir},\tilde{\rho}_{ir},\bar{\rho}_{ir} \leq 1 \quad (i=1,2,\dots,n)
		\end{aligned}
	\end{rcases}.
    \label{Optimization_eqn}
\end{equation}
Where, the global displacement vector is represented by $\mathbf{U}_r$, the vector $\mathbf{L}$ denotes the virtual load vector, in which all entries are zero except for the output degree of freedom, where the magnitude is set to $1$. The subscript $r$ corresponds to the robust design formulation, which accounts for both the blueprint and eroded configurations.  

The optimization objective is to minimize the maximum value between $\mathbf{L}^T\mathbf{U}_b$ and $\mathbf{L}^T\mathbf{U}_e$, where the subscripts $b$ and $e$ indicate the blueprint and eroded designs, respectively. These two terms basically represent the output displacements for the corresponding design. The matrices $\mathbf{K}_r$ and $\mathbf{T}$ represent the global stiffness and transformation matrices, while $\mathbf{F}_r$ and $\mathbf{P}_r$ denote the global force and pressure vectors, respectively.  

Here, $V_b$ and $V^*$ denote the volume of the blueprint design and the maximum allowable material volume within the structure, respectively. Similarly, $S_e$ and $S^*$ represent the strain energy of the eroded design and the permissible strain energy limit, respectively~\cite{kumar2024sorotop}. The scalar parameters $\Lambda_{b}$ and $\Lambda_{e}$, along with the column vectors $\bm{\lambda}_{1r}$ and $\bm{\lambda}_{2r}$, serve as the Lagrange multipliers associated with the respective constraints. The parameter $n$ denotes the total number of finite elements used to discretize the structure.

The Method of Moving Asymptotes (MMA)~\cite{svanberg1987method}, is employed here to update the design variables during the optimization process. Implementing this optimizer requires evaluating the sensitivities of both the objective function and its associated constraints. These sensitivities are computed using the adjoint-variable method. Further details on the sensitivity analysis procedure can be found in~\cite{kumar2024sorotop}.

\section{Results and discussions} \label{sec4}
\begin{figure}
	\centering
	\begin{subfigure}[b]{0.15\textwidth}
		\centering
		\includegraphics[scale=0.3]{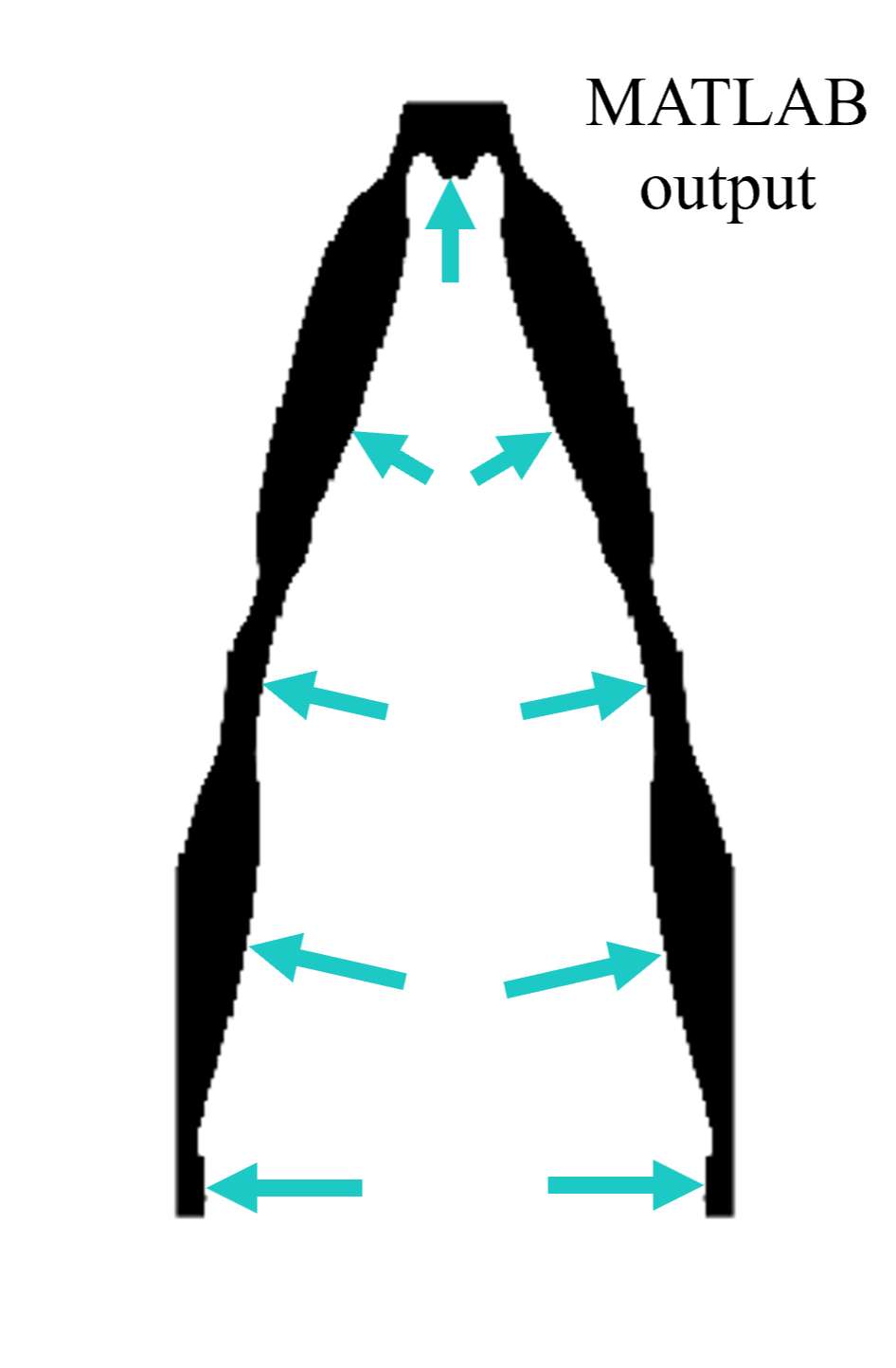}
		\caption{}
		\label{single_chamber_optimized_SPA}
	\end{subfigure}
    \quad
	\begin{subfigure}[b]{0.15\textwidth}
		\centering
		\includegraphics[scale=0.26]{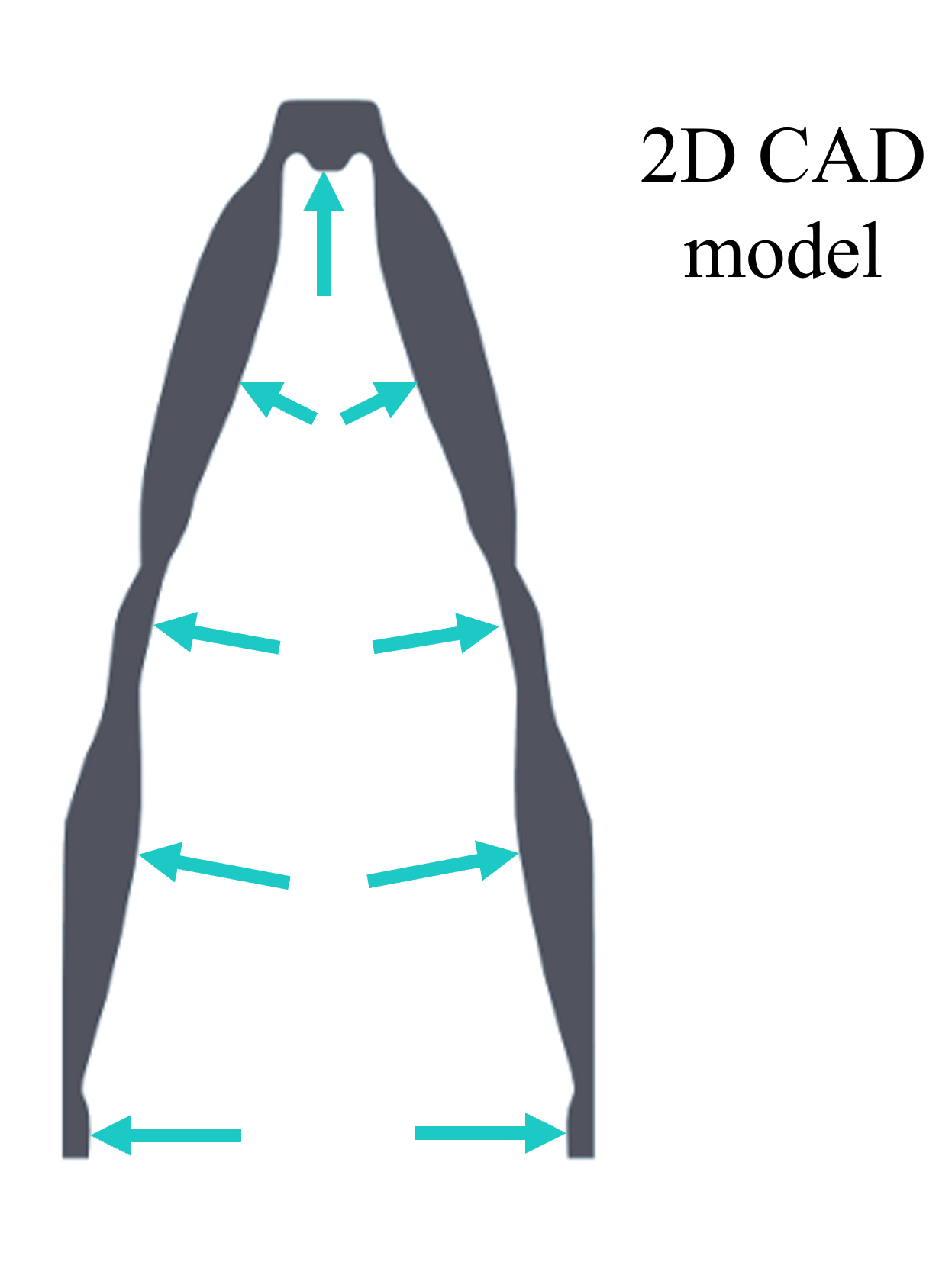}
		\caption{}
		\label{single_chamber_optimized_SPA_SWmodel}
	\end{subfigure}
    \begin{subfigure}[b]{0.15\textwidth}
		\centering
		\includegraphics[width=1.05\textwidth]
        {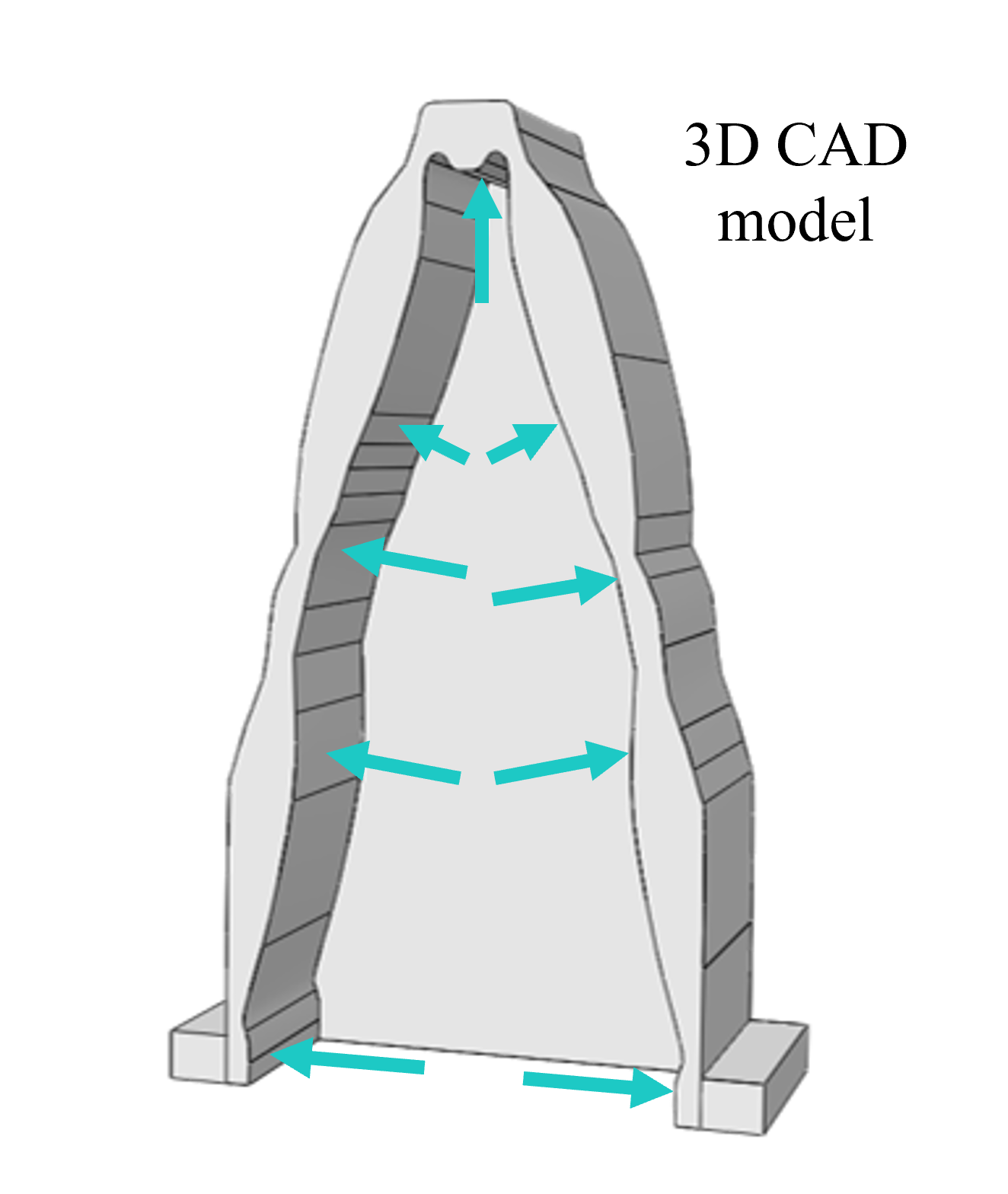}
		\caption{}
		\label{single_chamber_optimized_SPA_3d}
	\end{subfigure}
    \quad
	\caption{Optimized single SPA chamber with pressure load: (a) MATLAB-optimized output, (b) corresponding 2D CAD model, and (c) 3D CAD representation.}
	\label{optimized_SPA}
\end{figure}

\begin{figure}
	\centering
		\includegraphics[scale=0.7]
        {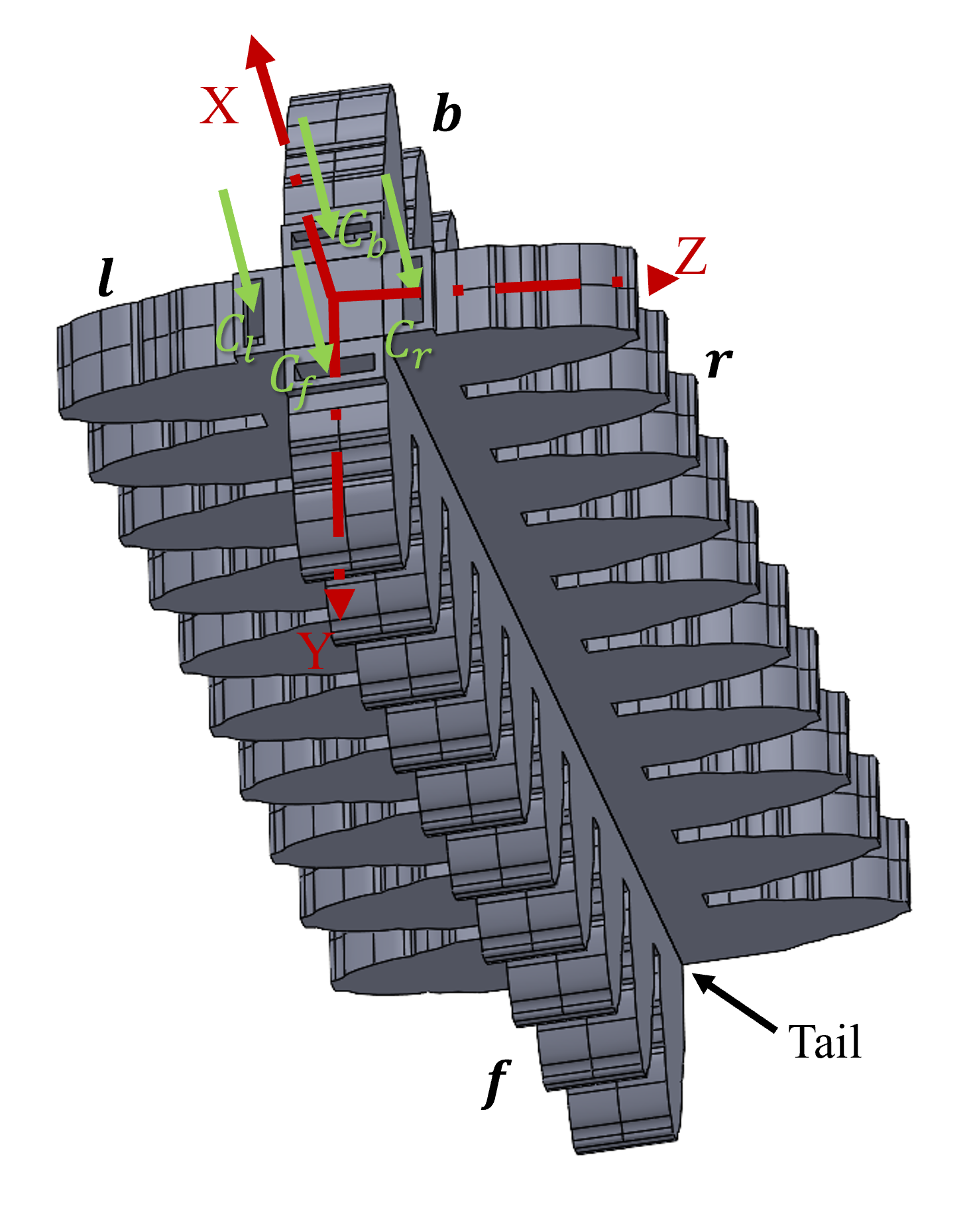}	
	\caption{3D Optimized SPA CAD model of an arm consisting of ten chambers in each of the four directions: front (\textbf{\textit{f}}), back (\textbf{\textit{b}}), left (\textbf{\textit{l}}), right (\textbf{\textit{r}}).}
	\label{optimized_SPA_3dview}
\end{figure} 

The symmetric half of the design domain (see Fig.~\ref{half_SPA}) is discretized in $80$ elements in the horizontal x-direction and in $320$  elements in the vertical y-direction using bilinear quadrilateral FEs. The maximum allowable volume fraction for the blueprint design is set to $0.20$, while the strain energy fraction associated with the eroded design is constrained to $80\%$. A filter radius of $7.6$ is used to ensure mesh independence and implicitly control the minimum member size. The flow parameters $\eta_k$ and $\beta_k$ are taken as 0.1 and 10, respectively. The same values are used for $\eta_d$ and $\beta_d$ to evaluate the Heaviside projection and the fluid flow fields. The stiffness of the workpiece spring is set as $k_{\text{o}} = 1$. The projection parameter $\beta$ is incrementally updated every 50 MMA iterations until reaching a value of 128, starting from an initial value of 1.

These parameters serve as inputs to the \texttt{SoRoTop} MATLAB code~\cite{kumar2024sorotop}, which generates the optimized design illustrated in Fig.~\ref{single_chamber_optimized_SPA}. The optimized chamber geometry notably differs from the conventional rectangular configuration shown in Fig.~\ref{full_SPA}.

Based on the optimized topology, a corresponding 2D CAD model is developed following the procedure described in~\cite{kumar2022topological}, as shown in Fig.~\ref{single_chamber_optimized_SPA_SWmodel}. The 2D CAD design is then extruded to form a 3D unit representing one chamber segment of the mdiSPA (Fig.~\ref{single_chamber_optimized_SPA_3d}). Multiple 3D units are sequentially patterned to construct the top symmetric section of a single arm, while the complete arm of the proposed mdiSPA is depicted in Fig.~\ref{optimized_SPA_3dview}.

\begin{figure}
    \centering
	\includegraphics[scale=0.8]
        {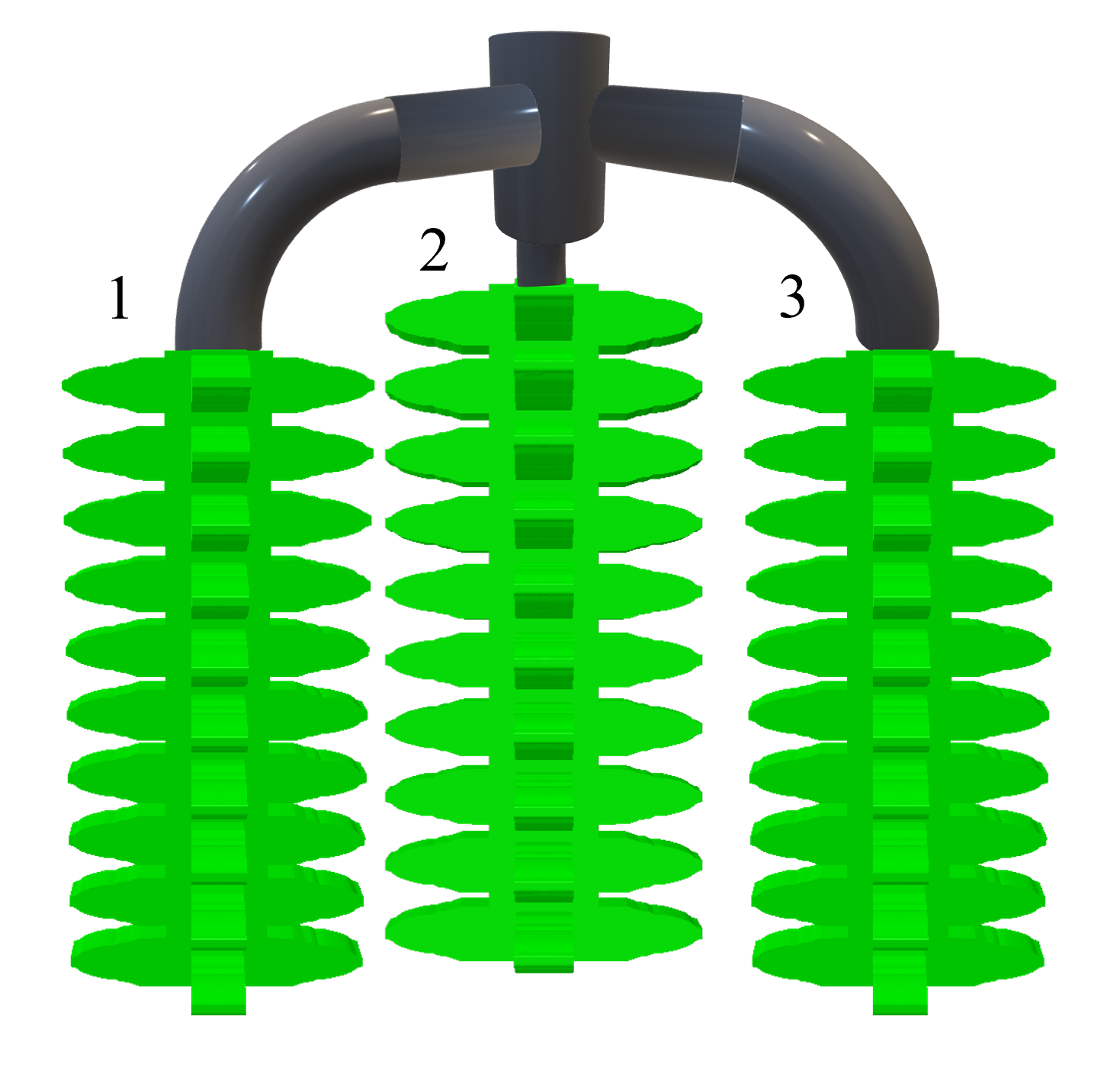}
    \caption{A representation of undeformed optimized SPA model to be used as a gripper.}
    \label{optimized_SPA_undeformed}
\end{figure}

\begin{figure}
    \centering
	\includegraphics[scale=0.7]
        {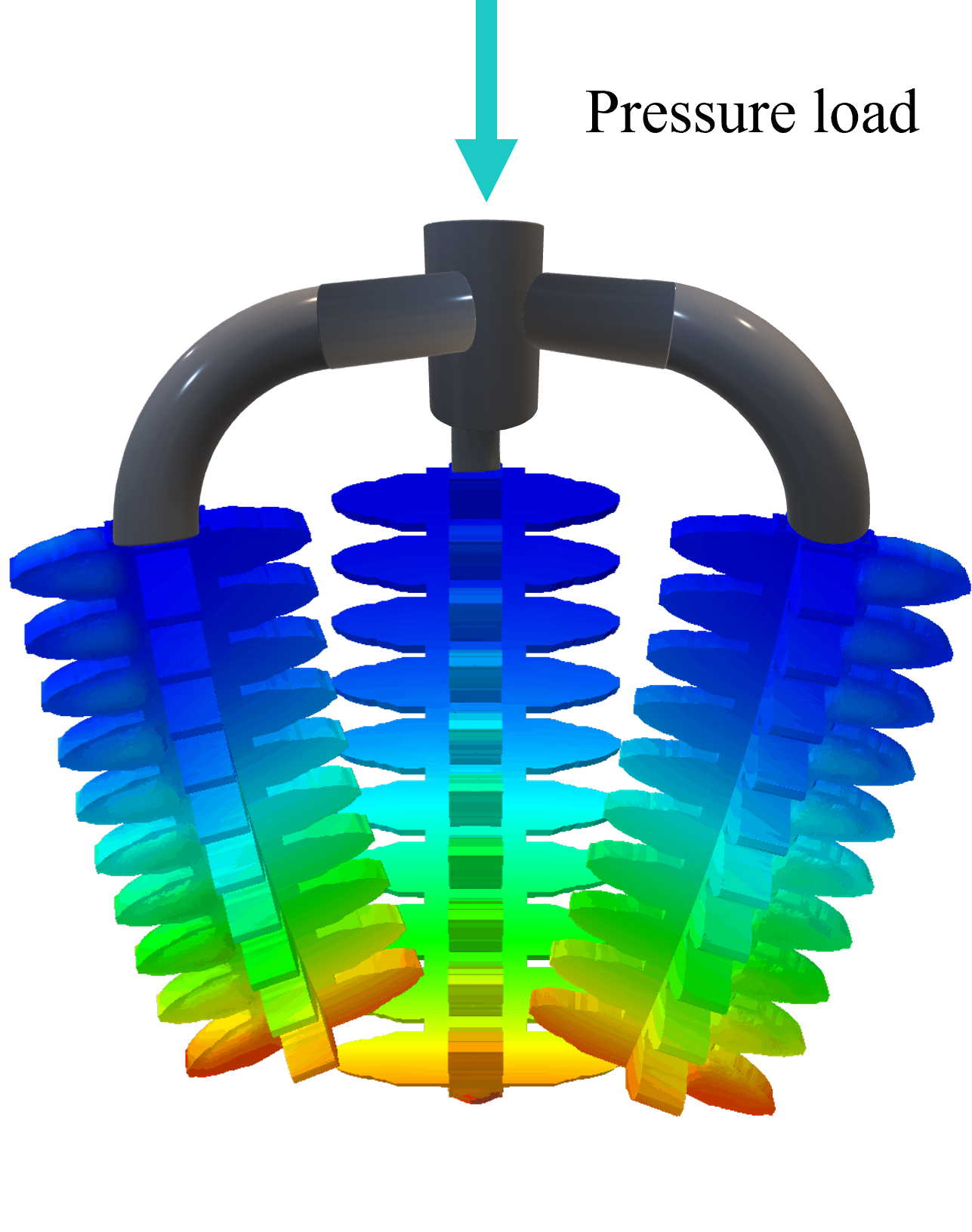}
    \caption{Deformed SPAs as a gripper - equal pressures are in front and left channels, back channel, and front and right channels in arm 1, arm 2, and in arm 3, respectively (Case 1).}
    \label{optimized_SPA_deformed_1}
\end{figure}

\begin{figure}
    \centering
    \includegraphics[scale=0.7]
        {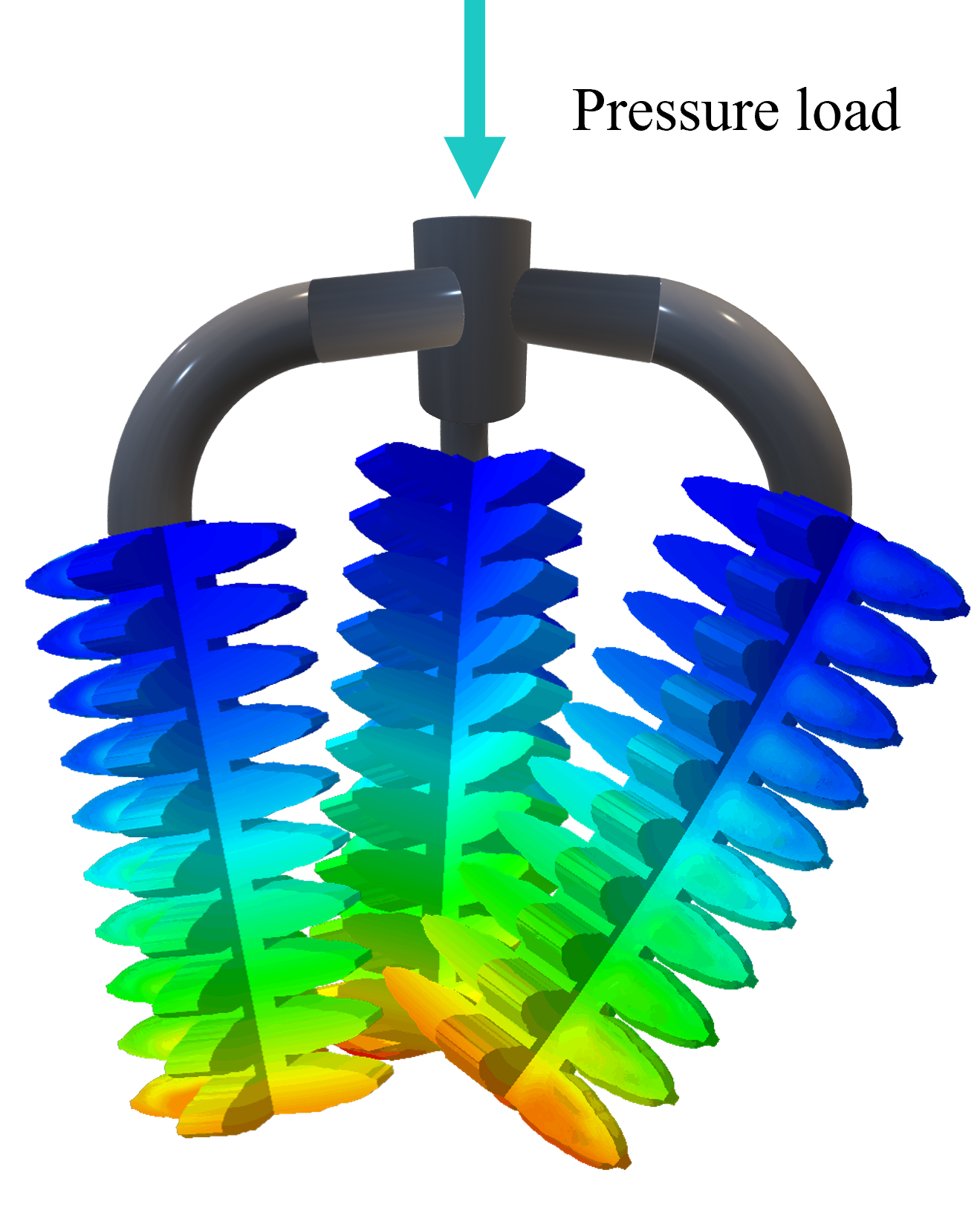}	
    \caption{Deformed SPAs as a gripper - different pressures are in front and left channels, front and right channels, and back and right channels in arm 1, arm 2, and in arm 3, respectively (Case 2).}
    \label{optimized_SPA_deformed_2}
\end{figure}

Figure~\ref{optimized_SPA_undeformed} illustrates the undeformed configurations of the optimized SPA, which consists of three interconnected arms, numbered as 1, 2, and 3, designed to function effectively as a gripper. Figure~\ref{optimized_SPA_deformed_1} shows the result when the pneumatic pressure is applied through channels $C_l$ and $C_f$ of arm 1, $C_b$ of arm 2, and $C_r$ and $C_f$ of arm 3. An equal amount of pressure is applied in these chambers. Figure~\ref{optimized_SPA_deformed_2} shows another result when the applied pressure loads are different. Here, the pressures are applied in channels $C_l$ and $C_f$ of arm 1, $C_b$ and $C_r$ of arm 2, and $C_r$ and $C_f$ of arm 3.

\begin{table}
\centering
\begin{tabular}{|ccccc|ccc|}
\hline
\multicolumn{5}{|c|}{Input pressure (MPa)} & \multicolumn{3}{|c|}{Deformations (mm)}\\ \hline
\multicolumn{1}{|c|}{\multirow{2}{*}{Arm}} & \multicolumn{4}{c|}{Channels} & \multicolumn{1}{|c|}{\multirow{2}{*}{$u_x$}} & \multicolumn{1}{|c|}{\multirow{2}{*}{$u_y$}} & \multicolumn{1}{|c|}{\multirow{2}{*}{$u_z$}}\\ \cline{2-5} 

\multicolumn{1}{|c|}{} & \multicolumn{1}{c|}{\textit{$C_f$}} & \multicolumn{1}{c|}{\textit{$C_b$}} & \multicolumn{1}{c|}{\textit{$C_l$}} & \multicolumn{1}{c|}{\textit{$C_r$}} & \multicolumn{1}{c|}{} & \multicolumn{1}{c|}{} & \multicolumn{1}{c|}{} \\ \hline

\multicolumn{1}{|c|}{1} & \multicolumn{1}{c|}{0.150} & \multicolumn{1}{c|}{-} & \multicolumn{1}{c|}{0.150} & \multicolumn{1}{c|}{-} & \multicolumn{1}{c|}{1.948} & \multicolumn{1}{c|}{-15.667} & \multicolumn{1}{c|}{15.658} \\ \hline

\multicolumn{1}{|c|}{2} & \multicolumn{1}{c|}{-} & \multicolumn{1}{c|}{0.150} & \multicolumn{1}{c|}{-}  & \multicolumn{1}{c|}{-} & \multicolumn{1}{c|}{0.994} & \multicolumn{1}{c|}{15.820} & \multicolumn{1}{c|}{0.004} \\ \hline

\multicolumn{1}{|c|}{3} & \multicolumn{1}{c|}{0.150} & \multicolumn{1}{c|}{-} & \multicolumn{1}{c|}{-}  & \multicolumn{1}{c|}{0.150} & \multicolumn{1}{c|}{1.948} & \multicolumn{1}{c|}{-15.667} & \multicolumn{1}{c|}{-15.658}\\ \hline
\end{tabular}
\caption{Case 1: Input and output data}
\label{Input_pressures1}
\end{table}

\begin{table}
\centering
\begin{tabular}{|ccccc|ccc|}
\hline
\multicolumn{5}{|c|}{Input pressure (MPa)} & \multicolumn{3}{|c|}{Deformations (mm)}\\ \hline
\multicolumn{1}{|c|}{\multirow{2}{*}{Arm}} & \multicolumn{4}{c|}{Channels} & \multicolumn{1}{|c|}{\multirow{2}{*}{$u_x$}} & \multicolumn{1}{|c|}{\multirow{2}{*}{$u_y$}} & \multicolumn{1}{|c|}{\multirow{2}{*}{$u_z$}}\\ \cline{2-5} 

\multicolumn{1}{|c|}{} & \multicolumn{1}{c|}{\textit{$C_f$}} & \multicolumn{1}{c|}{\textit{$C_b$}} & \multicolumn{1}{c|}{\textit{$C_l$}} & \multicolumn{1}{c|}{\textit{$C_r$}} & \multicolumn{1}{c|}{} & \multicolumn{1}{c|}{} & \multicolumn{1}{c|}{} \\ \hline

\multicolumn{1}{|c|}{1} & \multicolumn{1}{c|}{0.050} & \multicolumn{1}{c|}{-} & \multicolumn{1}{c|}{0.075} & \multicolumn{1}{c|}{-} & \multicolumn{1}{c|}{0.009} & \multicolumn{1}{c|}{-3.114} & \multicolumn{1}{c|}{5.212} \\ \hline

\multicolumn{1}{|c|}{2} & \multicolumn{1}{c|}{0.100} & \multicolumn{1}{c|}{-} & \multicolumn{1}{c|}{-}  & \multicolumn{1}{c|}{0.150} & \multicolumn{1}{c|}{1.131} & \multicolumn{1}{c|}{-7.809} & \multicolumn{1}{c|}{-15.751} \\ \hline

\multicolumn{1}{|c|}{3} & \multicolumn{1}{c|}{-} & \multicolumn{1}{c|}{0.080} & \multicolumn{1}{c|}{-}  & \multicolumn{1}{c|}{0.011} & \multicolumn{1}{c|}{0.031} & \multicolumn{1}{c|}{5.676} & \multicolumn{1}{c|}{-0.630}\\ \hline
\end{tabular}
\caption{Case 2: Input and output data}
\label{Input_pressures2}
\end{table}

The details of the input pressure loads and resulting data are shown in table \ref{Input_pressures1} and \ref{Input_pressures2}. The simulation is performed in Abaqus on the optimized CAD model. Isotropic ogden material model with $\mu_1=1.57$, $\alpha=2$, and $D_1=5\times10^{-5}$ is chosen. In the first case (case 1), the pressure is set to 0.015 MPa and applied through the corresponding channels. In arm 2, the pressure force is applied only to the backside chambers, and the deformation of the tail (the end point of the model on the central x-axis) in the y-direction is significantly greater than in the x- and z-directions. In the deformed configuration, the tail makes angles of 3.58\textdegree with the positive y-axis and 86.40\textdegree and 89.99\textdegree with the positive x- and z-axes, respectively. This shows that the tail has bent in the direction opposite to the chambers with applied pressures, with only a minute variation. In arm 1, the deformed tail makes angles of 84.97\textdegree, 134.80\textdegree, and 45.24\textdegree with the positive x-, y-, and z-axes, respectively. This is close to the resultant direction of the applied pressure force, as the equal pressure forces in the front and left channels produce a force at angles of 135\textdegree and 45\textdegree to the positive y- and z-axes, respectively. Similarly, we can see that in arm 3, the tail deforms in the resultant direction of the applied pressure. In case 2, pressures of 0.050 MPa and 0.075 MPa are applied through the front and left channels, respectively, in arm 1.
The tail of the arm 1 deforms at 89.92\textdegree, 120.86\textdegree, and 30.85\textdegree with the positive x-, y-, and z-axes, respectively. The resultant direction of the input pressure force is at 123.69\textdegree and 33.69\textdegree with the positive y- and z-axes, respectively. In arm 2, the applied pressures are 0.100 MPa and 0.150 MPa in the front and right chambers, respectively. The resultant will be at 146.31\textdegree with positive z-axis and 123.69\textdegree with positive y-axis. The actual deformation of the tail is at angles of 86.32\textdegree, 116.31\textdegree, and 153.39\textdegree with positive x-, y-, and z-axes, respectively. And finally in arm 3, the resultant of the applied pressure force is at an angle of 7.83\textdegree with the positive y-axis and 97.83\textdegree with the positive z-axis. At the same time, the deformed tail makes an angle of 6.35\textdegree and 96.33\textdegree with positive y- and z-axes, respectively. These results show that, with only input pressure values and the chambers where they are applied, one can predict the direction of the output deformation. That predicts the movement of one arm, and, when combined with the other arms of a gripper, it allows one to achieve the intended location and grip position. This utilizes the full potential of the arms to bend in multiple directions, enabling the gripper to hold in multiple orientations. The actuator demonstrates the bending behavior of each arm, which can be used together as a gripper. The incorporation of multiple arms in this mdiSPA configuration enables its effective use as a soft pneumatic gripper capable of adaptive and compliant grasping.

\section{CONCLUSIONS} \label{sec5}

The work presents a systematic design for constructing a multidirectional soft pneumatic actuator (mdiSPA) using topology optimization (TO). The proposed concept of utilizing the SPA in multiple directions to form a gripper using the TO approach is the core contribution of this work, enabling the development of a design that meets the desired functional requirements. The primary TO objective is to maximize the output displacement while constraining the strain energy within the eroded design to remain below a prescribed threshold and limiting the material volume of the blueprint design. The design-dependent nature of the pneumatic load is accurately modeled using Darcy’s law. The resulting optimized pressure chamber exhibits a distinctly different geometry compared to the conventional rectangular configuration. When subjected to pneumatic pressure, an actuator arm constructed from the optimized chamber exhibits significant bending. Utilizing this concept, a gripper mechanism is developed that can move and grip in multiple directions.

The developed mdiSPA achieves multidirectional bending motion under pneumatic actuation. This enables the use of mdiSPA in different applications, such as a gripper mechanism (as illustrated). A natural extension of this work for future research is the fabrication of the proposed mdiSPA via additive manufacturing and subsequent experimental validation to assess real-world performance.

\section*{Acknowledgment}
PK thanks the Anusandhan National Research Foundation, India, for support under the MATRICS project, file number MTR/2023/000524. 

%%%%%%%%%%%%%%%%%%%%%%%%%%%%%%%%%%%%%%%%%%%%%%%%%%%%%%%%%%%%%%%%%%%%%%%%%%%%%%%%

%\bibliographystyle{IEEEtran} 
%\bibliography{IEEEabrv,Citations}
% Generated by IEEEtran.bst, version: 1.14 (2015/08/26)

\end{document}